\documentclass[letterpaper, 10 pt, conference]{ieeeconf}
\IEEEoverridecommandlockouts
\usepackage{amsmath,amssymb,amsfonts}
\usepackage{algorithmic}
\usepackage{graphicx}
\usepackage{textcomp}
\usepackage{xcolor}
\usepackage{makecell}
\usepackage{multirow}
\usepackage{units}
\usepackage{subcaption}
\usepackage{censor}
\usepackage{float}

\usepackage[style=ieee,backend=biber,backref=false,hyperref=auto]{biblatex}
\newcommand{\quat}[1]{\boldsymbol{#1}}

\newcommand{\mymatrix}[1]{\boldsymbol{#1}}

\newcommand{\myvec}[1]{\boldsymbol{#1}}

\newcommand{\argmaximoneNoCon}[3]{\begin{aligned}#1\:  &  \underset{#2}{\arg\!\max}  &   &  #3
\end{aligned}
 }

\newcommand{\argminimfour}[7]{ \begin{aligned}#1\:  &  \underset{#2}{\arg\!\min}  &   &  #3 \\
  &  \text{subject to}  &   &  #4\\
  &   &   &  #5 \\
  &   &   &  #6 \\
  &   &   &  #7 \\
\end{aligned}
 }

\newcommand{\R}{\mathbb{R}}

\title{Coverage First Next Best View for Inspection of Cluttered Pipe Networks Using Mobile Manipulators}

\author{{Joshua R. Bettles,$^{1}$ Jiaxu Wu,$^{2}$ Bruno Vilhena Adorno,$^{1}$ Joaquin Carrasco,$^{1}$ and Atsushi Yamashita$^{2}$}%
\thanks{This work was partially supported by the ARTERD project as part of the OECD NEA NEST fellowship program 2025 and JAEA Nuclear Energy S\&T and Human Resource Development Project Grant Number JPJA24H24020026, EPSRC iCASE Project Grant Number EP/Y52864X/1, and by the Royal Academy of Engineering under the Research Chairs and Senior Research Fellowships programme.}%
\thanks{$^{1}$Joshua R. Bettles, Bruno Vilhena Adorno, and Joaquin Carrasco are with the Manchester Centre for Robotics and AI, The University of Manchester, Manchester, UK. Emails: { joshua.bettles@postgrad.manchester.ac.uk} and {\{bruno.adorno, joaquin.carrasco\}@manchester.ac.uk}.}%
\thanks{$^{2}$Jiaxu Wu and Atsushi Yamashita are with The University of Tokyo, Tokyo, Japan. Emails: {\{wujiaxu, yamashita\}@robot.t.u-tokyo.ac.jp}.}%
}

\begin{document}

\maketitle
 
\begin{abstract}
Robotic inspection of radioactive areas enables operators to be removed from hazardous environments; however, planning and operating in confined, cluttered environments remain challenging. These systems must autonomously reconstruct the unknown environment and cover its surfaces, whilst estimating and avoiding collisions with objects in the environment. In this paper, we propose a new planning approach based on next-best-view that enables simultaneous exploration and exploitation of the environment by reformulating the coverage path planning problem in terms of information gain. To handle obstacle avoidance under uncertainty, we extend the vector-field-inequalities framework to explicitly account for stochastic measurements of geometric primitives in the environment via chance constraints in a constrained optimal control law. The stochastic constraints were evaluated experimentally alongside the planner on a mobile manipulator in a confined environment to inspect a pipe network. These experiments demonstrate that the system can autonomously plan and execute inspection and coverage paths to reconstruct and fully cover the simplified pipe network. Moreover, the system successfully estimated geometric primitives online and avoided collisions during motion between viewpoints. 
\end{abstract}

\section{Introduction}

Pipe networks are ubiquitous across industries and particularly prevalent in the nuclear sector. Many nuclear power plants (NPPs) use pipes to transport effluent and steam across the site, and these networks require routine monitoring and inspection for damage and degradation. These pipe networks are located in radiological areas of NPPs, posing an additional health risk to workers who inspect them. These spaces are often confined and cluttered, making them difficult to monitor~\cite{OECD_Report}. Following natural disasters and incidents, such as the Fukushima Daiichi NPP incident, facilities may be damaged and therefore remain in an unknown physical and radiological state. 

These spaces require extensive professional monitoring to collect the data necessary to formulate an appropriate, targeted decommissioning plan, which exposes the expert to harmful radiation. Moreover, these experts are not always trained or qualified to perform structural monitoring tasks, necessitating secondary inspections, thereby increasing exposure to these hazardous environments. Mobile manipulators can navigate these confined spaces and collect data while mitigating the health and safety risks associated with the use of experts. The challenge is to generate inspection paths and control the mobile manipulator in these environments.

\subsection{Path Planning for Inspection Tasks}
Inspection of pipe networks can be categorised into two tasks: reconstruction and coverage. One approach for 3D volumetric reconstruction is next best view (NBV), which selects viewpoints to maximise the expected information gain (IG) of a sensor~\cite{Isler2016}. These approaches typically do not rely on prior knowledge of the environment and drive the system to explore the space. In~\cite{Bircher2016}, an RRT-based approach was used to generate and evaluate candidate branched paths within the environment. To address path generation in a computationally efficient way, a focal-point approach was utilised for a mobile manipulator to maintain focus on the object of interest when moving between viewpoints, rather than following branches~\cite{dursun2025objectreconstructionawarewholebodycontrolmobile}. A new approach to viewpoint generation was presented in~\cite{Vasquez-Gomez2014}, where viewpoints are generated within the configuration space of the mobile manipulator. The authors used a utility function to rank viewpoints based on collisions, overlap, discovery, and distance. A weighted sum of various IG metrics was introduced in \cite{Naazare2022}, based on unknown and visited voxels and sensor measurements from a radiation probe, where radiation sensor readings were encoded into a map a posteriori. 

Coverage path planning (CPP) is similar to NBV; however, it aims to generate paths that completely cover a known surface with a specific sensor. Many of these approaches sample candidate points in the vicinity of the object to be inspected and solve the set coverage problem using sampling-based methods, which usually provide probabilistic guarantees of complete coverage~\cite{Englot2013}. They can use both mesh- and occupancy-map-based representations of the object and optimise the path via resampling~\cite{Bircher2016_CPP}. In~\cite{Heng2015}, the problem of exploring and covering local unknown areas was considered. It was shown that, for increased path length, these approaches yield greater surface coverage. In \cite{Song2020}, an online global coverage and inspection planner was proposed that decomposes the space into sectors and inspects each region. The application in~\cite{Naazare2022} employs two separate sensors to inform coverage and exploration; however, it utilises the radiation sensor to inform the selection of the viewpoints for improved reconstruction of the target, as opposed to using the sensor to achieve a secondary task such as surface coverage.

\subsection{Collision Avoidance in Unknown Environments}

There are many approaches to collision avoidance in confined and cluttered environments, such as rapidly exploring random trees (RRTs), probabilistic roadmap (PRMs), artificial potential fields (APFs)~\cite{Batinovic2023}, and control barrier functions (CBFs)~\cite{Ames2019}. RRTs are exploration algorithms that navigate a space by generating a random tree that avoids collisions with obstacles in the environment~\cite{LaValle2006}. PRMs generate a sampling-based roadmap that captures the connectivity of the free configuration space; however, they typically handle static workspaces and require preprocessing to generate the roadmap~\cite{LaValle2006}. Also, since they are both planning approaches, they are used in addition to a controller. APFs are widely used in planning and control applications; however, they suffer from oscillations, local minima, and unreachable targets~\cite{Batinovic2023}. Conversely, CBFs enforce set invariance on the controller's input, ensuring that the system remains within the safe set \cite{Ames2019}. They have been applied across the control field for collision avoidance, most notably in adaptive cruise control~\cite{Ames2014_cruise}, bipedal robotic walking~\cite{Hsu2015_bipedal}, and manipulation~\cite{Ferraguti2020_manipulation}.

Vector field inequalities (VFIs) \cite{Marinho2019} can be considered a special case of CBFs~\cite[remark 4]{Ames2019}. The authors adopted a geometric approach to derive task-space collision-avoidance constraints that, by construction, generate the forward-invariant safe set $\Omega_S$~\cite{Marinho2019}. In \cite{dursun_maintaining_2023}, these were implemented alongside a whole-body kinematic controller for a holonomic mobile manipulator to avoid obstacles and maintain visibility of a moving object. VFIs have been extended to second-order systems in~\cite{Quiroz-Omana2019} and applied to self-collision avoidance between elements of the robotic system. In~\cite{Marinho2022}, an adaptive controller was derived to adapt system parameters online and account for kinematic inaccuracies in the model. In that work, VFIs were modified to enforce collision avoidance under the control input and adaptation signals in the estimated task space.

Those controllers and constraints are formulated in continuous time, assuming deterministic knowledge of the geometric primitives within the environment; as such, they only provide formal safety guarantees when operating in continuous time with perfect knowledge of the constraints. Derwent et al. \cite{Derwent2025} addressed the discretisation error by buffering the constraints to ensure that system behaviour remained safe despite the discrete implementation of continuous time formulations. The CBF field addresses this problem by redesigning the controller in discrete time using discrete-time Control Lyapunov Functions and discrete-time CBFs~\cite{Agrawal2017, Xiong2023}. Several works have extended CBFs to stochastic systems, with \cite{Clark2021} formulating reciprocal and zero CBFs for continuous-time stochastic systems. Meanwhile, \cite{Lin2026} investigated stochastic CBF under state estimation for systems that evolve on manifolds and demonstrated the approach for systems evolving in $SE(2)$ and $SE(3)$. In \cite{Mestres2025}, a probabilistic CBF framework for discrete-time stochastic systems was developed by reformulating the CBF as chance constraints. Chance constraints have also been widely used in stochastic model predictive control~\cite{Farina2016}, with most research focusing on the effects of uncertainty and noise in the state variable. 

\subsection{Statement of Contributions}
The contributions of this paper are twofold. Firstly, we propose a new planning scheme that enables simultaneous execution of coverage and inspection tasks using two independent sensors rather than a single sensor as in current state-of-the-art approaches. Recasting the CPP problem as a coverage IG problem enables integration with current NBV approaches without loss of generality and facilitates its application to other tasks with differing sensor models and requirements. We apply this to radiation monitoring in an unknown environment with no prior radiological measurements, and to reconstructing surfaces within a confined space.

Secondly, this paper introduces the concept of stochastic VFIs (SVFIs), which extends the current VFI framework to account for measurement uncertainty in obstacles. By introducing chance constraints into the constrained optimal motion controller, we account for uncertainty in the geometric primitives, thereby providing collision avoidance with probabilistic guarantees. We transform these stochastic constraints via a deterministic surrogate, which can be easily included in the current constrained optimal controller formulations.

Both approaches were experimentally validated on a nonholonomic mobile manipulator for the task of inspecting an unknown pipe network in a confined space to simulate a real-world environment.

\section{Mathematical Background}
\subsection{Whole-Body Kinematic Control}
The task-space kinematics of a mobile manipulator can be described as
\begin{align}
    \myvec{x} &= \myvec{h}(\myvec{q}),\label{FK}\\
    \dot{\myvec{x}} &= \mymatrix{J}\myvec{u},\label{DKE}
\end{align}
where $\myvec{x}\in \R^m$ is the task-space vector, $\myvec{q}\triangleq \myvec{q}(t) = [\myvec{q}_{\mathrm{base}}^T\quad \myvec{q}_{\mathrm{arm}}^T]^T\in \mathbb{R}^n$ is the stacked configuration vector of the robot's base and arm, $\myvec{h}:\R^n\to\R^m$ is a function that maps the configuration vector to the task-space vector, typically obtained through the forward kinematics, $\mymatrix{J}\triangleq\mymatrix{J}(\myvec{q})\in\mathbb{R}^{m\times n}$ is the task Jacobian matrix, and $\myvec{u}$ is the control input given as configuration velocities (i.e., $\myvec{u}\triangleq\dot{\myvec{q}}$). 

Let the task be described as $\myvec{x} \triangleq [\myvec{t}_e^T \quad \myvec{n}_e^T]^T$, in which $\myvec{t}_e, \myvec{n}_e\in\mathbb{R}^3$ is the translation and direction (i.e., rotation axis) of the end effector, respectively. The task Jacobian is given as $\mymatrix{J} \triangleq [\mymatrix{J}_t^T\quad \mymatrix{J}_n ^T]^T$, where $\mymatrix{J}_t\triangleq \mymatrix{J}_t(\myvec{q})$ is the translation Jacobian and $\mymatrix{J}_n\triangleq \mymatrix{J}_n(\myvec{q})$ is the line direction Jacobian that satisfies $\dot{\myvec{n}}_e=\mymatrix{J}_n \dot{\myvec{q}}$ ~\cite{Marinho2019}.

The desired whole-body control law is formulated as a constrained quadratic program (QP)~\cite{Marinho2019} that enforces the task-space error to converge to zero exponentially as best as possible. It is given by
\begin{equation}\label{eq:controller}
    \argminimfour{\myvec{u}\in}
    {\dot{\myvec{q}}}
    {\left|\left|\mymatrix{J}\dot{\myvec{q}}+\kappa{\Tilde{\myvec{x}}}\right|\right|^2_2 + \lambda_c\left|\left|\dot{\myvec{q}}\right|\right|^2_2  + \lambda_s||\myvec{s}||^2_2}
    {\mymatrix{W}_{i}\dot{\myvec{q}}\preceq \mymatrix{w}_{i}}{ \mymatrix{W}_{e}\dot{\myvec{q}}= \mymatrix{w}_{e}}
    {\mymatrix{W}_{s}\dot{\myvec{q}} - \myvec{s} \preceq \mymatrix{w}_{s}}
    {\Pr(\mymatrix{W}_{c}\dot{\myvec{q}}\succeq \mymatrix{w}_{c})\succeq \myvec{\alpha}_c,}
\end{equation}
where $ \myvec{u}$ is the control input expressed as configuration space velocities, $\Tilde{\myvec{x}} = \myvec{x} - \myvec{x}_d$ is the task-space error with controller gain and damping given by $\kappa,\lambda_c\in(0,\infty)$, respectively. The set point is assumed to be constant such that $\dot{\myvec{x}}_d = \myvec{0}$. The system is subject to a set of linear differential inequalities in the control inputs, in which $\mymatrix{W}_{i}$ and $\myvec{w}_{i}$  enforce task-space and configuration-space constraints through deterministic VFIs. The nonholonomic constraint is enforced by $\mymatrix{W}_{e}$ and $ \myvec{w}_{e}$ as an equality constraint on the QP using the method of~\cite{Quiroz-Omana2017}. The whole-body control law is the same as that in~\cite {dursun2025objectreconstructionawarewholebodycontrolmobile}, which was used for a holonomic system. Specific deterministic VFIs with parameters given by $\mymatrix{W}_{s}$ and $\myvec{w}_{s}$ are relaxed through the use of the slack vector $\myvec{s}$ to ensure feasibility~\cite{Derwent2025}, which is penalised by $\lambda_s\in(0,\infty)$. We propose additional chance constraints to ensure that the VFIs with parameters given by $\mymatrix{W}_{c}$ and $\myvec{w}_{c}$, related to uncertain geometric primitives, will be respected with a probability of at least $\myvec{\alpha}_c$. 

\subsection{Vector Field Inequalities}
VFIs are a geometrical approach to CBF that enables the transformation of nonlinear constraints in the robot configuration into linear differential inequalities in the robot configuration. They are particularly useful to enforce collision avoidance between two geometric primitives, where one or more are kinematically coupled to the robot. They leverage the signed distance between the primitives and the distance Jacobian that relates the time derivative of the distance to the robot's configuration velocities. For dynamic objects, it is important to determine the residual, the motion of the entities that is independent of robot configuration velocities. In this present work, however, only static objects are being considered. The reader is directed to~\cite{Marinho2019} for the derivation of the constraints, residuals, and implementation. For completeness, the essentials are summarised herein.

VFIs can be used to enforce two behaviours, staying within a safe zone or outside of a restricted zone with a (constant) minimum safety distance $d_{\mathrm{safe}} \in [0,\infty)$. The distance to the boundary is given by $\Tilde{d}\triangleq\Tilde{d}(\myvec{q}) = d - d_{\mathrm{safe}} $, where $d\triangleq d(\myvec{q})$ is the distance between two primitives in the task-space, with one being kinematically coupled to the robot. The restricted zone $\Omega_R$ and the safe zone $\Omega_S$ given as
\begin{align*}
\Omega_R &\triangleq \left\{ \myvec{q} \in \mathbb{R}^n:\, \Tilde{d}(\myvec{q}) < 0 \right\},
\\
\Omega_S &\triangleq \left\{ \myvec{q} \in \mathbb{R}^n:\, \Tilde{d}(\myvec{q}) \geq 0 \right\}.
\end{align*}

The signed distance dynamics can be expressed as 
\begin{equation}
    \dot{\tilde{d}} = \dot{d} - \dot{d}_{\mathrm{safe}}=\dot{d},
\end{equation}
where a positive $ \dot{\tilde{d}}$ means the robot is moving away from the boundary, whereas a negative $ \dot{\tilde{d}}$ means the robot is approaching the boundary. These dynamics are constrained by 
\begin{equation}\label{eq:vfi_dynamics_joint_velocities}
     \dot{\tilde{d}} \geq -\eta \tilde{d} \iff  -\mymatrix{J}_d\dot{\myvec{q}} \leq \eta\Tilde{d},
\end{equation}
where $\mymatrix{J}_d\in\R^{1\times n}$ is the distance Jacobian relating the primitive attached to the robot to the workspace primitive, and $\eta \in [0,\infty)$ is the maximum allowable approach rate to the boundary.

\section{Stochastic Vector Field Inequalities for Planes}
We introduce SVFIs in \eqref{eq:controller} to account for measurement uncertainty by exploiting chance constraints. We consider only the uncertainty associated with estimating geometric primitives in the environment. Since the geometrical parameters of these primitives are not known exactly, the associated covariances must be accounted for in the controller to prevent collisions with a prescribed probability. 

A chance constraint imposes a probability that a constraint is satisfied and allows the inclusion of uncertainties within the formulation. The general form is given by 
\begin{equation}\label{eq:chance}
    \Pr\left(\mymatrix{A}\myvec{x} \succeq \myvec{b}\right) \geq \alpha,
\end{equation}
where $\alpha \in (0,1)$ is the probability that the constraint is satisfied, $\mathbb{R}^n\ni \myvec{x}\sim \mathcal{N}(\myvec{\mu}_x, \mymatrix{\Sigma}_x)$, $\mathbb{R}^m \ni\myvec{b} \sim \mathcal{N}(\myvec{\mu}_b, \mymatrix{\Sigma}_b)$ and $\mathbb{R}^{m \times n}\ni\mymatrix{A} \sim \mathcal{N}(\myvec{\mu}_A, \mymatrix{\Sigma}_A)$. We use \eqref{eq:vfi_dynamics_joint_velocities}  in \eqref{eq:chance} to obtain
\begin{gather}\label{eq:chance_dynamics_joint}
    \Pr\left(\dot{\tilde{d}} \geq -\eta \tilde{d}\right) \geq \alpha\iff \Pr\left(\mymatrix{J}_d\dot{\myvec{q}} \geq -\eta\Tilde{d}\right) \geq \alpha.
\end{gather}

Let a static plane be given by $\myvec{\pi} = [\myvec{n}_\pi^T\quad d_\pi]^T$, where $\myvec{n}_\pi\in\mathbb{R}^3$ is the normal of the plane, and $d_\pi\in\R$ is the signed perpendicular distance between the plane and the origin of the reference frame. The estimation of the true plane is given by $\Hat{\myvec{\pi}}   \sim \mathcal{N}(\myvec{\pi}, \mymatrix{\Sigma}_{\pi})$, with $\Hat{\myvec{\pi}} = [\Hat{\myvec{n}}_\pi^T\quad \Hat{d}_\pi]^T$, and $\mymatrix{\Sigma}_{\pi} \in \mathbb{R}^{4\times4}$ being the covariance matrix associated with the plane estimation. Given an arbitrary point $\myvec{t}\triangleq\myvec{t}(\myvec{q})$ attached to the robot, with $\dot{\myvec{t}}=\mymatrix{J}_t\dot{\myvec{q}}$, the point-to-plane distance with respect to $\myvec{\pi}$ is given by 
\begin{equation}\label{eq:point-to-plane-distance}
    d^\pi_{t,\pi} = \langle \myvec{n}_\pi, \myvec{t}\rangle - d_\pi.
\end{equation}

The  desired distance dynamics, enforced through the use of VFIs  \cite{Marinho2019},  is expressed as
\begin{equation}\label{eq:desired_distance_dynamics}
\dot{d}^\pi_{t,\pi}\geq-\eta\tilde{d}^\pi_{t,\pi}\iff \myvec{n}_\pi^{T}\mymatrix{J}_{t}\dot{\myvec{q}}\geq-\eta(d^\pi_{t,\pi}-d_{\mathrm{safe}}),
\end{equation}
which implies $\myvec{n}_\pi^{T}\mymatrix{J}_{t}\dot{\myvec{q}}+\eta d^\pi_{t,\pi}\geq\eta d_{\mathrm{safe}}$. By substituting \eqref{eq:point-to-plane-distance} into the VFI \eqref{eq:desired_distance_dynamics}, we obtain
\begin{equation}
    \myvec{n}_\pi^T\left(\mymatrix{J}_{t}\dot{\myvec{q}}+\eta\myvec{t}\right) - \eta d_\pi \geq\eta d_{\mathrm{safe}}.
\end{equation}

By defining
\begin{gather}
f(\myvec{\pi}) \triangleq \myvec{n}_\pi^T\left(\mymatrix{J}_t\dot{\myvec{q}} + \eta{\myvec{t}}\right) - \eta d_\pi= \begin{bmatrix}
\dot{\myvec{q}}^{T}\mymatrix{J}_{t}^{T}\!+\!\eta \myvec{t}^{T} & -\eta\end{bmatrix}
\myvec{\pi}
\label{eq:function_f}
\end{gather}
as a linear function of the plane parameters and substituting it into \eqref{eq:chance_dynamics_joint}, we get
\begin{equation}\label{eq:plane_chance_constraint}
    \Pr\left(f({\myvec{\pi}}\right)\geq\eta d_{\mathrm{safe}})\geq \alpha,
\end{equation}
which expresses that the VFI will be respected with probability at least $\alpha$. Assuming that the estimated plane parameters follow a Gaussian distribution, we can express the function \eqref{eq:function_f} in terms of the estimated plane as $f(\Hat{\myvec{\pi}})\sim \mathcal{N}(f(\myvec{\pi}), \sigma_{f({\myvec{\pi}})}^2)$. The covariance of $f(\myvec{\pi})$ is propagated through the linear transformation~\cite{Gu2021} given by 
\begin{gather}
\sigma_{f({\myvec{\pi}})}^2 = \nabla f\left({\myvec{\pi}}\right)^T \myvec{\Sigma}_{\pi} \nabla f\left({\myvec{\pi}}\right),
\label{eq:covariance_propagation}
\end{gather}
where
\begin{gather}
    \nabla f({\myvec{\pi}}) = 
        \begin{bmatrix}
            \left(\frac{\partial f({\myvec{\pi}})}{\partial \myvec{n}_\pi}\right)^T \\
            \frac{\partial f({\myvec{\pi}})}{\partial {d_\pi}}
        \end{bmatrix}  =   
        \begin{bmatrix}
            \mymatrix{J}_t\dot{\myvec{q}} + \eta{\myvec{t}} \\
            -\eta
        \end{bmatrix}.
        \label{eq:gradient_of_f}
\end{gather}

We define $Z \triangleq \frac{f(\myvec{\pi}) - {f(\myvec{\Hat{\pi}})}}{\sigma_{f(\myvec{{\pi}})}} \sim \mathcal{N}(0,1)$  and use \eqref{eq:covariance_propagation}  to rewrite \eqref{eq:plane_chance_constraint} as 
\begin{align}\label{eq:Pr_Z}
     \Pr\left(Z \geq \frac{\eta d_{\mathrm{safe}} - {f(\myvec{\Hat{\pi}})}}{\sigma_{f(\myvec{{\pi}})}}\right) = 1 - \Phi\left(\frac{\eta d_{\mathrm{safe}} - f(\myvec{\Hat{\pi}})}{\sigma_{f(\quat{\pi})}}\right) &\geq \alpha,
\end{align}
where $\Phi$  is the standard Gaussian cumulative density function \cite{Papoulis2002}. This allows us to use the inverse cumulative distribution function, $\Phi^{-1}$, on \eqref{eq:Pr_Z}.
 Noting that $\Phi(-x)=1-\Phi(x)$ and that $\Phi$ is a monotonically increasing continuous function, we obtain
\begin{equation*}
    \Phi\!\left(\!\frac{  f(\myvec{\Hat{\pi}})\!-\!\eta d_{\mathrm{safe}}}{\sigma_{f(\quat{\pi})}}\!\right) \!\geq \!\alpha\iff \eta d_{\mathrm{safe}} -{f(\myvec{\Hat{\pi}})} \!\leq\! -\Phi^{-1}(\alpha)\sigma_{f(\myvec{{\pi}})}.
\end{equation*}

The final deterministic surrogate of the SVFI is given as 
\begin{equation}
    - \underbrace{ \Hat{\myvec{n}}_\pi^T\mymatrix{J}_{t}}_{ {\mymatrix{J}}_{\Hat{\pi}}}\dot{\myvec{q}} \leq \eta( \underbrace{\Hat{\myvec{n}}_\pi^T{\myvec{t}} - \Hat{d}_\pi}_{{d}_{\Hat{\pi}}} - d_{\mathrm{safe}}) - b_{\alpha},
    \label{eq:deterministic_surrogate_VFI}
\end{equation}
where $b_{\alpha} = \Phi^{-1}(\alpha)\sigma_{f({\myvec{\pi}})}$, ${\mymatrix{J}}_{\Hat{\pi}}$ is the expected-point-to-plane distance Jacobian, and ${d}_{\myvec{\Hat{\pi}}}$ is the expected-point-to-plane distance.

The term $b_{\alpha}$ in  \eqref{eq:deterministic_surrogate_VFI}  serves to buffer the point-to-plane VFI based on the covariance of the geometric parameters of the measured plane and the chosen probability of respecting the VFI, thereby giving the buffer value a probabilistic meaning, whilst being generalisable to other geometric primitives.

The constraint \eqref{eq:deterministic_surrogate_VFI}  is nonlinear in the control inputs ($\dot{\myvec{q}}$) since the covariance depends on joint velocities. In practical discrete implementations based on zero-order hold, the configuration velocities calculated in the latest sampling period can be used in the calculation of the covariance of the plane's geometric parameters. However, this will affect the buffer size, especially during acceleration towards the buffer. Since $|b_{\alpha} | \propto \|\dot{\myvec{q}}\|$, if the configuration velocity vector norm from the previous sampling period used to calculate the buffer is smaller than the actual velocity in the current sampling period, the buffer size might be underestimated (overly confident estimate). This may result in a violation of the inflated buffer and, consequently, the VFI might not be respected with the prescribed probability. Conversely, if the configuration velocity vector is larger in the previous sampling period than in the actual sampling period, the buffer will be overestimated due to an overly conservative estimate. Nonetheless, those issues are unlikely to have any practical effect for sufficiently small sampling periods. 

\section{Coverage First Next Best View}

To plan multiple tasks simultaneously, such as 3D reconstruction and surface coverage, we present our coverage first next best view (CFNBV) planner. It reformulates each task using IG and a linearly weighted sum, which is then applied to two independent sensors. This approach is generalisable to any number of tasks or sensors, provided that an appropriate sensor model and an IG metric are specified.

The IG in CFNBV is defined as the amount of information each candidate viewpoint provides about the environment per task. The concept of volumetric information (VI) is defined in \cite{Isler2016}, which uses a voxelised representation of the environment to quantify the IG based on voxel occupancy. We use the VI framework to define visual IG and entropy, and apply these concepts to CPP as coverage IG and entropy. 

Let us consider the bounded voxelised robot workspace $\mathcal{W}=\bigcup_{i=1}^{N_\nu} {\nu}_i$, where ${\nu}_i\subset \R^3$ is the $i$th voxel and $N_\nu$ is the total number of voxels, containing unknown surfaces such as pipes, discretised and represented as an OctoMap \cite{octomap}. Initially, the workspace is unknown,  and the goal of the visual portion (3D reconstruction) of the algorithm is to categorise each voxel such that $\mathcal{W} = \mathcal{W}_{\mathrm{unk}}\cup \mathcal{W}_{\mathrm{free}}\cup \mathcal{W}_{\mathrm{occ}} \cup \mathcal{W}_{\mathrm{res}} $, where  $\mathcal{W}_{\mathrm{unk}}$ is the set of unknown voxels, $\mathcal{W}_{\mathrm{free}}$ is the set of unoccupied voxels,  $\mathcal{W}_{\mathrm{occ}}$ is the set of occupied voxels, and $\mathcal{W}_{\mathrm{res}}$  is the set of residual voxels. A residual voxel is one where there exists no valid viewpoint from which to observe it; this occurs for voxels inside or obstructed by objects. The goal is to then drive the system to achieve $\mathcal{W}_{\mathrm{unk}} = \varnothing$, where the equality becomes $\mathcal{W}\setminus \mathcal{W}_{\mathrm{res}} = \mathcal{W}_{\mathrm{free}}\cup \mathcal{W}_{\mathrm{occ}}$. Conversely, the goal of the coverage portion of the algorithm is to achieve $\mathcal{W}_{\mathrm{cov}}=\mathcal{W}_{\mathrm{occ}}$, where $\mathcal{W}_{\mathrm{cov}}$ is the set of covered voxels, by categorising each voxel based on a secondary sensor. The task is considered complete only when both conditions are satisfied.

\subsection{Visual Information Gain}
The rear side entropy from \cite{Isler2016} is used to define visual IG. For a given candidate viewpoint in $\mathcal{V} \ni\myvec{v} = [\myvec{t}_c^T \quad \myvec{n}_c^T]^T$, where $\myvec{t}_c,\myvec{n}_c\in\R^3$ are the candidate voxel position and randomly generated direction, respectively, and a visual sensor model, let $\mathcal{R}_v(\myvec{v})$ denote the set of random rays cast from the viewpoint within the sensor model. We thus define the expected IG from the candidate voxel viewpoint $\mathcal{G}_v({\myvec{v}})$ to be
\begin{equation}
    \mathcal{G}_v({\myvec{v}}) \!=\!\!\! \sum_{r\in\mathcal{R}_v(\myvec{v})}\!\mathcal{I}_{\mathcal{R}_v}(r), \text{ with } \mathcal{I}_{\mathcal{R}_v}(r)=  \sum_{\nu\in \mathcal{W}_{v}}\mathcal{I}_v(\nu),
\end{equation}
where $\mathcal{I}_{\mathcal{R}_v}(r)$ is the IG for each ray $r$ and $\mathcal{I}_v(\nu)$ is the VI for each voxel $\nu\in\mathcal{W}_{v}\subseteq \mathcal{W}$ that the ray $r$ passes through until it reaches a surface or the maximum sensor range. The VI of $\nu$ is given by the entropy of the voxel, defined as 
\begin{equation}\label{eq:visual_VI}
\mathcal{I}_o(\nu) = -\Pr(\nu)\ln{(\Pr(\nu))} - \overline{\Pr}(\nu)\ln{(\overline{\Pr}(\nu))},
\end{equation}
where $\overline{\Pr}(\nu) = (1-\Pr(\nu))$ and $\Pr(\nu)$ is the probability that the voxel $\nu$ is occupied. We can also define a binary function based on the observed state of each voxel as 
\begin{equation}\label{eq:binary_visual}
    \mathcal{I}_p(\nu) = \begin{cases}
        1, & \text{if }\nu \text{ is unknown},\\
        0, & \text{if }\nu \text{ is known},
    \end{cases}
\end{equation}
such that combining \eqref{eq:visual_VI} and \eqref{eq:binary_visual} gives the expression
\begin{equation}
    \mathcal{I}_v(\nu) = \mathcal{I}_p(\nu) \mathcal{I}_o(\nu),
\end{equation}
where the IG is only for unobserved voxels.

\subsection{Coverage Information gain}
Since we are only interested in occupied voxels when considering the coverage task, we cannot use the same definitions as for the visual VI gain to define the coverage VI gain. Given a different sensor model, such as a sphere representing a radiation probe that is to cover a surface, define $\mathcal{R}_o(\myvec{v})$ as the set of rays cast from $\myvec{v}$ within the coverage sensor model. Let us also consider a new state for each voxel as ``covered", which denotes whether a voxel has been covered or observed by this secondary sensor. The coverage IG, $\mathcal{G}_{c}({\myvec{v}})$, is given as
\begin{equation}
    \mathcal{G}_c({\myvec{v}}) \!=\!\!\! \sum_{r\in\mathcal{R}_o(\myvec{v})}\!\mathcal{I}_{\mathcal{R}_o}(r), \text{ with } \mathcal{I}_{\mathcal{R}_o}(r)=  \sum_{\nu\in \mathcal{W}_\mathrm{o}}\mathcal{I}_c(\nu),
\end{equation}
where $\mathcal{I}_{\mathcal{R}_o}(r)$ is the IG for each ray $r$ and $\mathcal{I}_c(\nu)$ is the VI for each voxel $\nu\in\mathcal{W}_{o}\subseteq \mathcal{W}$ that the ray $r$ passes through until it reaches a surface or the maximum sensor range.
Let us define the coverage VI, $\mathcal{I}_c(\nu)$, for each voxel as a binary function based on the covered and occupied states as 
\begin{equation}\label{eq:cover_info}
    \mathcal{I}_{c}(\nu)\! =\! \begin{cases}
        \ln(0.5), \!\!\!\!& \text{if } \nu\text{ is occupied and not covered},\\
        0,                 \!\!\!\!& \text{if } \nu \text{ is unoccupied, unknown, or covered}.
    \end{cases}
\end{equation}

\subsection{Weighted Information Gain}
The two IGs are combined via a weighted linear sum, with weights determined by the task priority; any number of tasks may be considered. For our application, the coverage and visual IG are combined as
\begin{equation}
    \mathcal{G}_w(\myvec{v}) = \beta\mathcal{G}_v(\myvec{v}) + (1-\beta)\mathcal{G}_c(\myvec{v})
\end{equation}
where $\beta\in [0,1]$ is the task priority and can be altered in favour of specific tasks. 

\subsection{Sampling and Selection}
Uniform random sampling is used to generate the set of candidate viewpoints at each planning step. Suppose we have a function $\mathrm{Uniform}: 2^\mathcal{W} \to \R^3\times\R^3$ that generates a random candidate viewpoint from a set of free voxels, where $2^\mathcal{W}$ is the power set of $\mathcal{W}$. We define the set $\mathcal{V}_N$,  containing $N$ samples, such that $\mathcal{V}_N = \{\myvec{v}_i\in\R^3\times\R^3:\myvec{v}_i=\mathrm{Uniform}(\mathcal{W}_{\mathrm{free}}), i = 1,\ldots, N \}$. 
To find the optimal candidate viewpoint at each planning step, it suffices to solve
\begin{equation}
    \argmaximoneNoCon{\myvec{x}_d\in}{\myvec{v} \in\mathcal{V}}{\mathcal{G}_w(\myvec{v}),}
\end{equation}
where $\myvec{x}_d$ becomes the controller's new set point. The process repeats once the set point is reached, the error norm of the closed-loop system formed of \eqref{FK} and \eqref{DKE} under control law \eqref{eq:controller} reaches a local minimum, or the stopping criterion is met.

\begin{table}[ht]
    \vspace{2mm}
    \caption{Experimental Parameters}\label{tab:experiment_parameters}
    \centering\renewcommand{\arraystretch}{1.15}
    \begin{tabular}{l l c}
        \hline
        & \textbf{Parameter} & \textbf{Value} \\
        \hline\hline
    
        \multirow{3}{*}{\makecell[l]{\textbf{Sensor}\\\textbf{Configuration}}} & RGB-D field of view (FoV) & $(60^\circ, 45^\circ)$ \\
        & RGB-D maximum range & $3\,\mathrm{m}$ \\
        & Radiation sensor radius & $0.4\,\mathrm{m}$ \\
        \hline
    
        \multirow{5}{*}{\makecell[l]{\textbf{Controller}\\\textbf{Parameters}}} & $(\alpha, \kappa, \lambda_c)$& $(0.7, 6, 1.2)$\\
        & $(\lambda_s, \eta)$ & $( 5\cdot 10^3, 1)$\\
        & Point-to-plane $d_\mathrm{safe}$ (arm, base)   & $\unit[(0.1,0.5)]{m}$\\
        & Point-to-line $d_\mathrm{safe}$  & $0.075\,\mathrm{m}$\\
        & Frequency & $100\,\mathrm{Hz}$ \\
        \hline
        
        \multirow{5}{*}{\makecell[l]{\textbf{Planning}\\\textbf{Parameters}}} & No. of rays in $\mathcal{R}_o$, $\mathcal{R}_v$ & $200$ \\
        & $N$ & $500$ \\
        & Octomap resolution & $0.05\,\mathrm{m}$ \\
        & Stopping criterion & $\mathrm{IG} < 1$ \\
        & Error norm criterion & $||\Tilde{\myvec{x}}||_2 <  0.001$ \\
        & $\beta$ & $0.75$ \\
        \hline
    \end{tabular}
\end{table}
\section{Experimental Setup}
The system comprises a UF7 manipulator and an AgileX Tracer nonholonomic mobile platform with an onboard power distribution system and a wireless router. An on-board computer with AMD Ryzen 7 8-core CPU running at 3.8 GHz and 32 GB of RAM was used, running Ubuntu 22.04 and ROS~2 Humble. The drivers for each robot and the whole-body controller were implemented using the SmartArmStack~\cite{aiscienceplatform2024} and the DQ Robotics library~\cite{Adorno2021DQControl}, respectively. Localisation was handled via FAST-LIO~\cite{xu2021fastliofastrobustlidarinertial}, and the map was stored as an OctoMap~\cite{octomap} for planning. Geometric primitives were extracted using RANSAC from the point cloud library~\cite{Rusu2011}. An external laptop with an Intel Core i7 8-core CPU running at 3.8 GHz and 16 GB of RAM running Ubuntu 22.04 was connected to the robot via Wi-Fi and used to visualise the output and execute the CFNBV planner to generate inspection points. An Intel RealSense D455 was used to capture visual data for updating the OctoMap of the environment.

\begin{figure}[t]
    \centering
    \includegraphics[width=0.8\linewidth]{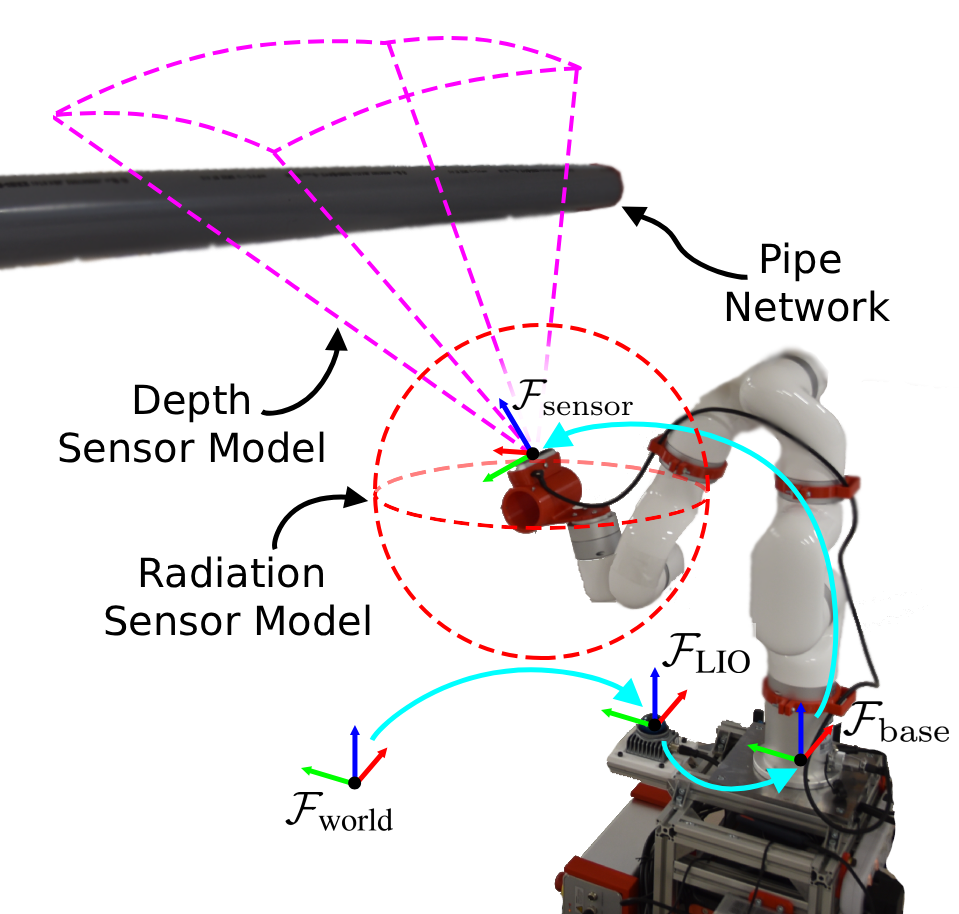}
    \caption{The experimental setup with the constructed mobile manipulator in a single-pipe environment, overlaid with the radiation sensor model represented as a sphere in red, centred on $\mathcal{F}_\mathrm{sensor}$, and the depth sensor model represented by the cameras' FoV in pink.}
    \label{fig:system}
\end{figure}

The experiment was confined to a $1.5\,\mathrm{m}$ cube with an arbitrary configuration of one to three pipes. Self-collision avoidance was implemented to ensure that the sensors did not collide with the arm or base~\cite{Quiroz-Omana2019}. A slack variable was used for the joint velocity and point-to-line constraints~\cite{Marinho2019} to prevent the control problem from becoming infeasible by allowing minor violations to compensate for noisy measurements and the line estimator's poor performance. The experimental parameters of the system are given in Table~\ref{tab:experiment_parameters}. A depth sensor model was used to estimate visual IG, as shown in Fig.~\ref{fig:system}, and a radiation sensor model was used to estimate coverage IG.

Several experimental limitations include the absence of a radiation sensor. Thus, a spherical sensor model was used, with random data generated to represent background radiation sensor readings. Since the planner does not use the radiation sensor measurements and only considers the sensor model, the absence of an actual sensor does not affect its performance. The selected depth camera had a minimum distance requirement of $30\,\mathrm{cm}$; erroneous readings occurred when the system moved too close to a pipe. These erroneous readings occasionally caused the pipe estimator to fail, resulting in collisions between the robot and the pipe network. This could be mitigated by using a more appropriate depth sensor with a minimum distance better suited to the close interaction with the pipe required by the radiation sensor.
\section{Results and Discussions}
A total of nine successful trials were conducted with the system in an unknown environment with varying pipe configurations. This section focuses on the results from the three-pipe scenario, as it is the most challenging configuration tested.

\subsection{Controller}
\begin{figure*}[t]
    \centering
    \subfloat[]{\includegraphics[width=0.31\textwidth]{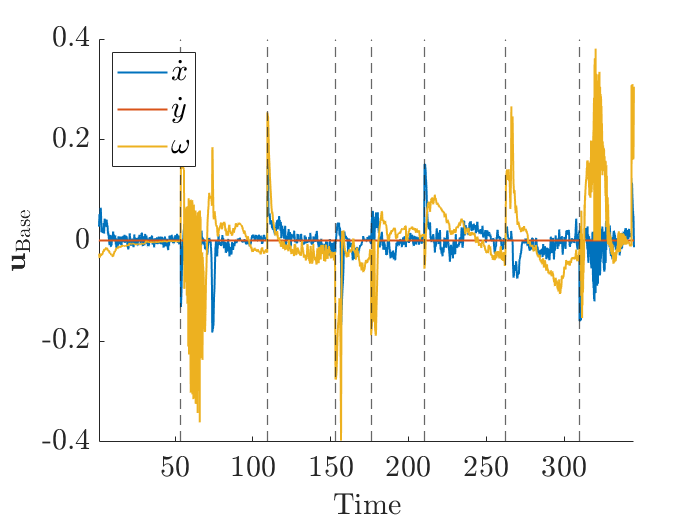}
    \label{fig:base_velocities}}
    \subfloat[]{\includegraphics[width=0.31\textwidth]{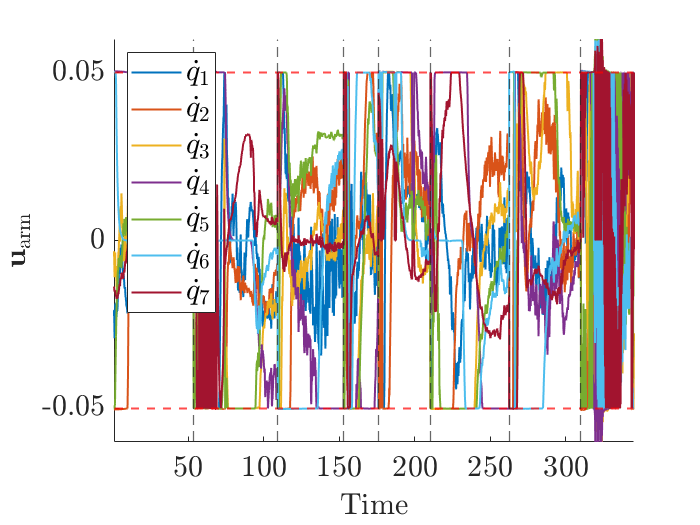}
    \label{fig:joint_velocities}}
    \subfloat[]{\includegraphics[width=0.31\textwidth]{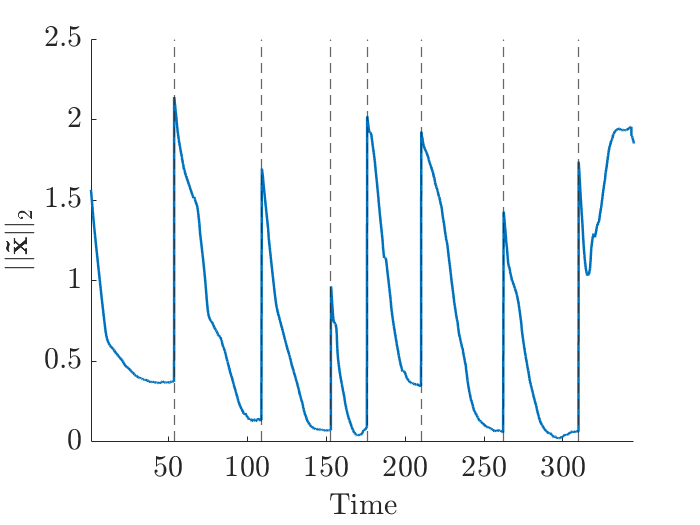}
    \label{fig:task_space_error}}
    \vspace{0.03cm}
    \subfloat[]{\includegraphics[width=0.31\textwidth]{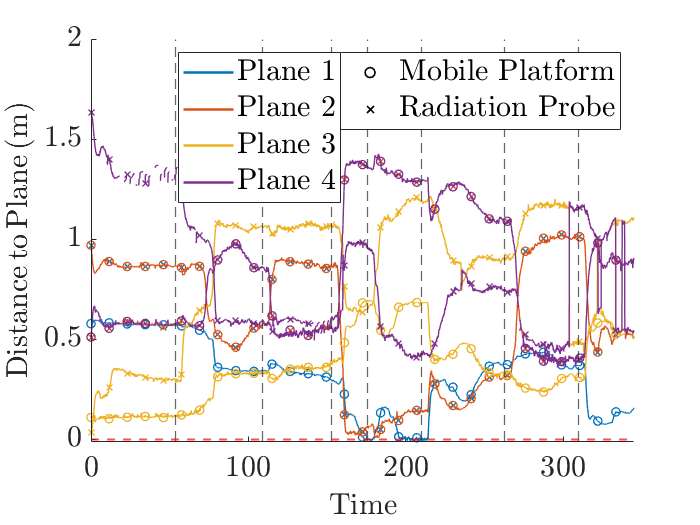}
    \label{fig:plane_constraints}}
    \subfloat[]{\includegraphics[width=0.31\textwidth]{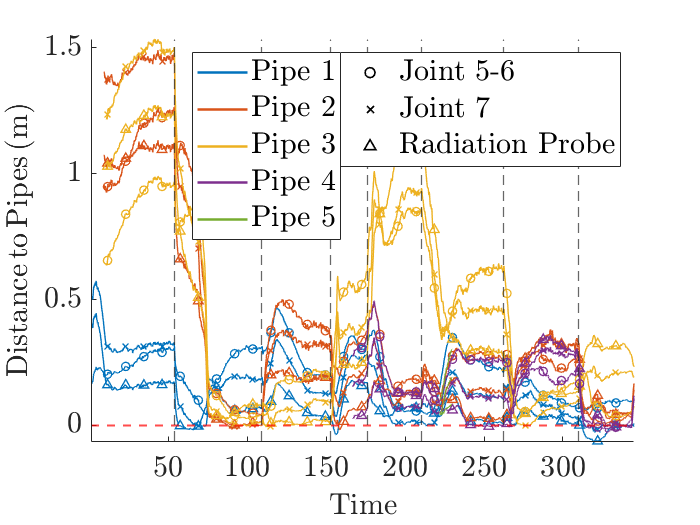}
    \label{fig:pipe_constraints}}

    \caption{Time response of the closed-loop system from one trial in the three-pipe environment: (a) control input for the base, (b) control input for the arm, (c) norm of the task-space error, (d) distance to planes, and (e) distance to pipe constraints for the mobile manipulator in the confined space. Horizontal red dashed lines represent limits for the associated variables, whereas vertical black dashed lines represent the selection of a new set point.}
    \label{fig:robot_results}
\end{figure*}

The whole-body controller \eqref{eq:controller} was implemented, and the time response is shown in Fig.~\ref{fig:task_space_error} for one of the trials with three pipes. The control input for the base, Fig.~\ref{fig:base_velocities}, expressed in the base frame, shows $\dot{y} = \unitfrac[0]{m}{s}$ for the duration of the trial, meaning the nonholonomic equality constraint was always satisfied, as expected.

During all trials, the expected-point-to-plane constraints were satisfied for the mobile platform and the radiation probe, where the underlying distance constraint was maintained even as the buffer suffered minor violations, as shown in Fig.~\ref{fig:plane_constraints}. All joint limits were respected throughout the experiments. 

Each spike in the task-space error was due to the controller selecting a new target, as shown in Fig.~\ref{fig:task_space_error}, which occurred when the error norm remained unchanged for more than four seconds or fell below the threshold. At certain points during execution, the task-space error increased (at $175$ and $315$ seconds in Fig.~\ref{fig:task_space_error}). This was due to violations of the unbuffered, deterministic point-to-line constraints, which caused the system to increase the distance to the target to reach a safe state and navigate around the boundary. This caused the control input for the arm joint velocities to exceed the limit in order to force the system out of the restricted area, Fig.~\ref{fig:joint_velocities}. However, these are only minor violations and plateau due to the introduction of the slack variable and its heavy penalisation.

\subsection{CFNBV}
\begin{figure*}[t]
    \centering
    \subfloat[]{\includegraphics[width=0.31\textwidth]{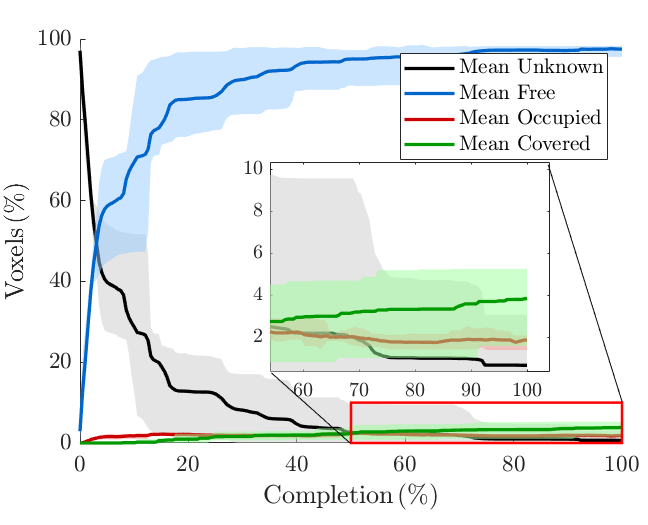}
    \label{fig:Voxels}}
    \subfloat[]{\includegraphics[width=0.31\textwidth]{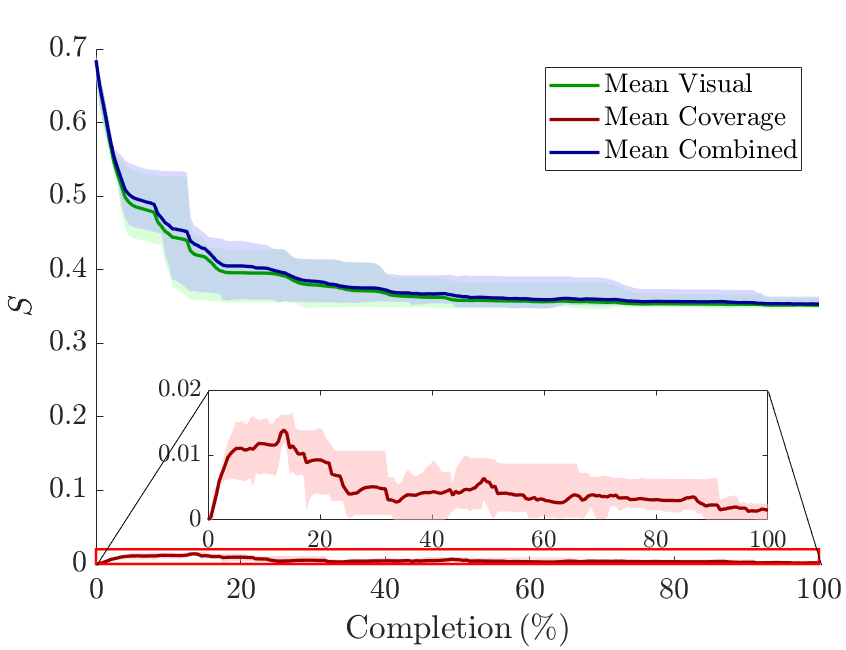}
    \label{fig:Entropy}}
    \subfloat[]{\includegraphics[width=0.31\textwidth]{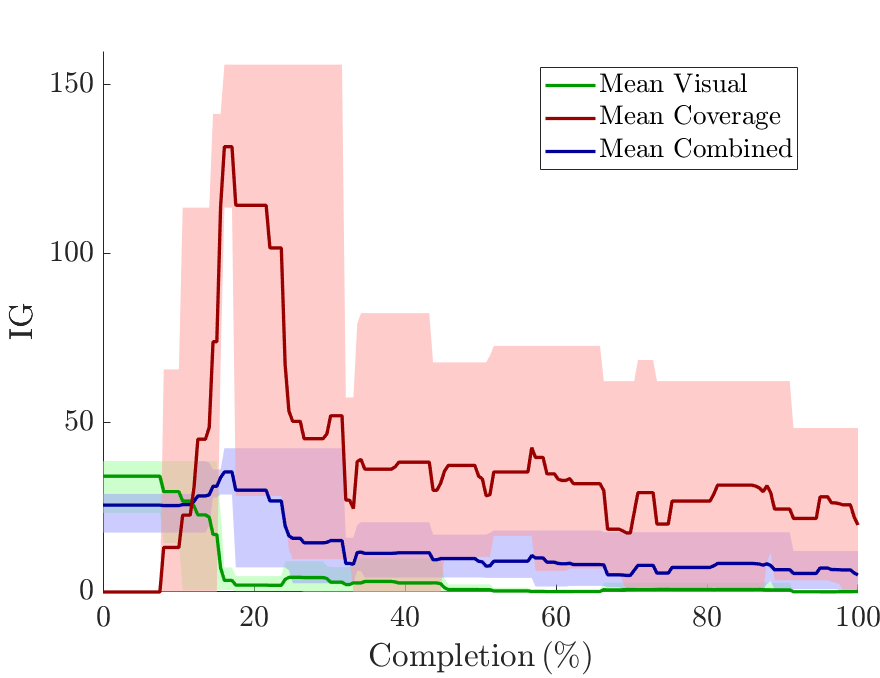}
    \label{fig:IG}}

    \caption{Results of the CFNBV planner within the three-pipe environment with the mean value of the five trials displayed with the maximum and minimum values overlaid. (a) The evolution of the voxels and their categorisation based on the visual and coverage sensors, (b) the minimisation of the combined entropy per voxel, and (c) the interplay between the IGs for environment exploration and surface exploitation.}
    \label{fig:planner}
\end{figure*}

Five of the nine trials were within the three-pipe environment, with the system starting at a location where no pipes were visible. The initial viewpoint selected maximises visual IG, as shown in Fig.~\ref{fig:IG}, resulting in an exponential decrease in the percentage of unknown voxels and average entropy within the first $20\%$ of the task as the closed-loop system converges to the initial set point. 

During this time period, the number of both the free and occupied voxels increased as the pipes within the environment were detected, shown in Fig.~\ref{fig:Voxels}. This resulted in an increase in the coverage entropy per voxel, as new information is introduced into the system, which can be seen in Fig.~\ref{fig:Entropy}. The mean combined entropy per voxel of the system is defined using \eqref{eq:visual_VI} and \eqref{eq:cover_info} as $S = \frac{1}{|\mathcal{W}|}\sum_{ \nu\in \mathcal{W}}\big(\mathcal{I}_o(\nu) + \mathcal{I}_{c}(\nu)\big)$, where $|\mathcal{W}|$ is the cardinality of $\mathcal{W}$. 

This increase in entropy is reflected in the interplay between the visual and coverage IG: once a surface has been identified, the optimisation automatically switches from exploring the environment to exploiting the surface, balancing coverage of uncovered voxels whilst choosing viewpoints that still generate visual information. This leads to a locally optimal selection of viewpoints at each planning iteration and to an approach that aims to minimise entropy and the information of the system. 

This approach is a probabilistic implementation of NBV due to the random generation of candidate viewpoints. The position was generated by selecting a voxel within $\mathcal{W}_\mathrm{free}$ according to a uniform distribution, whereas the viewpoint axis direction was generated by selecting pan and tilt angles from a uniform distribution in the intervals $[-\pi,\pi)$ and $[-\frac{\pi}{2},\frac{\pi}{2}]$, respectively. 

Since each voxel in the OctoMap is represented by a probability of occupancy, these values approach but never reach the occupancy limits, and therefore, the visual entropy of the system will never decrease to zero. Conversely, since the coverage entropy is represented in a binary fashion, this will decrease to zero if all voxels are covered. 

The current system does not reclassify occupied and covered voxels if they later become unoccupied due to depth sensor noise, leading to misclassification. This resulted in a covered voxel percentage exceeding that of occupied voxels. We chose not to reassess the coverage of reclassified voxels, as it would potentially require unnecessary secondary coverage of these points.

\section{Conclusions}
This paper introduces a new planning method for inspecting and covering unknown surfaces in confined spaces. It reformulates the CPP problem as a coverage IG problem and integrates it with NBV approaches. By defining a suitable information-gain metric for the sensor and application, tasks can be combined into a single planning approach, rather than necessitating multiple planners. The performance of this approach was experimentally evaluated on a mobile manipulator in a confined environment, demonstrating the utility of simultaneous reconstruction and coverage.

Moreover, the VFI framework was extended to account for measurement uncertainty in the geometric primitives. Estimating these primitives online removes the need for prior knowledge of the environment and for precise localisation with respect to the robot. By incorporating uncertainty, we provide a probability of collision avoidance for these estimated primitives and show that it is equivalent to adding a buffer to a surrogate deterministic constraint, which can be dynamically calculated online.

The proposed method requires an estimator to extract the pose of each primitive; here, it is applied to a plane. However, generalising this to other primitives is the topic of ongoing work.
The current approach also does not account for uncertainty in the mobile manipulator's configuration or in its actuators, whose effects must still be modelled and incorporated into both the controller and the constraints. Future work for the planner shall demonstrate and validate its versatility across alternative platforms and tasks, whilst addressing voxel reclassification and providing formal guarantees on coverage and global behaviour.

\printbibliography[title={References},heading=bibintoc]

\end{document}